%% file: main.tex
\documentclass{bmvc2k}

\title{ Turbo Training with Token Dropout}
\addauthor{Tengda Han}{htd@robots.ox.ac.uk}{1}
\addauthor{Weidi Xie}{weidi@robots.ox.ac.uk}{1,2}
\addauthor{Andrew Zisserman}{az@robots.ox.ac.uk}{1}

\addinstitution{
 Visual Geometry Group\\
 Department of Engineering Science\\
 University of Oxford\\
}
\addinstitution{
Coop.\ Medianet Innovation Center \\ Shanghai Jiao Tong University \\ Shanghai, China
}

\runninghead{Han, Xie, Zisserman}{Turbo Training with Token Dropout}

\input{sec/def}
\newcommand{\app}[0]{{Appendix}}

\usepackage{caption}
\definecolor{bmvc_blue}{RGB}{0,0,102} 
\usepackage{bm}
\usepackage{comment}
\usepackage{bbding}
\usepackage{comment}
\usepackage{amssymb}
\usepackage{multirow}
\usepackage{booktabs}
\usepackage{subfigure}
\usepackage{todonotes}
\usepackage{graphicx}
\usepackage[font={small}]{caption}
\usepackage{comment}
\usepackage{tabu}

\begin{document}

\maketitle

\input{sec/01-abstract}
\input{sec/02-introduction}
\input{sec/03-related_work}

\input{sec/04-method}

\input{sec/05-experiment}
\input{sec/06-result}
\input{sec/07-conclusion}

\bibliography{bib/shortstrings,bib/vgg_local,bib/vgg_other}

\clearpage
\vspace{10mm}
\section*{{\LARGE Appendix}}
\vspace{5mm}
\appendix
\input{sec/08-supp}

\end{document}

%% file: sec/def.tex
\usepackage{pifont}
\newcommand{\cmark}{\ding{51}}%
\newcommand{\xmark}{\ding{55}}%

\usepackage{xpunctuate}

%% file: sec/01-abstract.tex
\begin{abstract}
The objective of this paper is an efficient training method for video tasks.
We make three contributions:
(1) We propose Turbo training,
a simple and versatile training paradigm for Transformers 
on multiple video tasks.
(2) We illustrate the advantages of Turbo training on
action classification, video-language representation learning,
and long-video activity classification, showing that
Turbo training can largely maintain competitive performance while
achieving almost $4\times$ speed-up and significantly less memory consumption.
(3) Turbo training enables long-schedule video-language training and 
end-to-end long-video training, 
delivering competitive or superior performance than previous works,
which were infeasible to train under limited resources.
\end{abstract}

%% file: sec/02-introduction.tex
\section{Introduction}
\label{sec:intro}
\vspace{-2mm}
The extra temporal axis of video, compared to a static image, can capture rich information such as long term
activities and stories. However, it also brings a few orders of magnitude of more data with the concomitant
costs in processing and memory requirements. This, together with 
the long training schedules required to train networks on video data, has tremendously slowed research progress in video understanding and has raised the bar for researchers to make contributions in this area, particularly with
limited resources. Furthermore, training consumes a significant amount of energy and hardware, with detrimental consequences for the natural environment.

In this paper, 
we propose a computation-efficient paradigm for training Transformers on video tasks, termed as \textbf{Turbo training}, 
which only needs to process sparsely sampled visual tokens. 
It is based on a multi-task training that jointly optimises two objectives:
a masked region autoencoder loss, 
and the standard training loss for the downstream task of interest, 
for example, cross-entropy for action recognition, or contrastive loss for visual-language representation learning.

Turbo training is possible because the visual data contains redundancy. 
As observed in the recent self-supervised pre-training work, such as MAE~\cite{He2022MAE} for images 
and~\cite{Tong2022VideoMAE,Feichtenhofer2022VideoMAE} for videos, 
it is sufficient to use only a small fraction of the input visual tokens
to reconstruct the visual signal.
For example, in images, 75\% of the visual tokens can be dropped, 
whilst in videos, the dropout can be even more aggressive, and 90\% of the spatio-temporal tokens can be dropped while still being able to learn a strong representation using self-supervision.
In Turbo training, we adopt a similar intuition and target  efficient supervised training for video understanding.
In short:  
\textbf{given the high redundancy in videos, 
training Transformers with sparsely sampled visual tokens is sufficient for the downstream tasks of interest}. 
This is beneficial for video understanding tasks in two ways:
{\em first, training efficiency}, 
it dramatically reduces the computational cost for training transformers; 
{\em second, memory requirements}, videos of longer length can be loaded and directly trained on, leading to improved performance and new abilities,
{\em i.e.}, end-to-end training the transformer models. 

To summarise, we make the following contributions: 
(1) We propose Turbo training, 
a new paradigm for training Transformers on video understanding tasks, 
that jointly optimises a masked autoencoding and standard training loss 
for the downstream task of interest. 
(2) We illustrate the advantages of Turbo training on three
video understanding tasks, including action classification, 
video-language training, and long-video activity classification,
showing that training on sparsely sampled video patches can 
largely maintain competitive performance on most of the video tasks of interest, 
while achieving almost $4\times$ speed-up and significant reduction in memory consumption.
As a consequence, 
(3) Turbo training enables 
long-schedule video-language representation learning and
end-to-end long-video activity classification,
delivering competitive or superior performance than previous methods.

%% file: sec/03-related_work.tex
\vspace{-5mm}
\section{Related Works}
\label{sec:related_works}
\vspace{-1mm}

\paragraph{Masked Autoencoding}
has been used in the natural language processing community with great success to pre-train language models with a fill-in-the-blank task;
specifically, the model is trained to use the context words surrounding a masked token to try to predict what the masked word should be, a typical example would be BERT~\cite{Devlin18}. 
In fact, a similar idea has also been investigated in computer vision for self-supervised learning in the seminal work of context autoencoder~\cite{Pathak16}, where the authors proposed  to learn a visual representation by training a ConvNet to inpaint the missing region of an image. However, due to the large effective spatial receptive fields of ConvNets, 
severe boundary artefacts weakened the learnt representation.
Recently, with the advent of the Vision Transformers~\cite{Dosovitskiy20}, that do not suffer from the same spatial receptive field problem, 
masked autoencoding has become popular again 
as an effective method to pre-train large-scale Vision Transformers. 
Examples include MAE~\cite{He2022MAE}, SimMiM~\cite{Xie22SimMiM}, and BEiT~\cite{Bao22BEiT} for images, and~\cite{Feichtenhofer2022VideoMAE,Tong2022VideoMAE,Gupta22maskvit} for videos.\\[-0.8cm]

\paragraph{Efficient Deep Neural Networks.}
Since the resurgence of deep neural networks, 
reducing the computational cost for training or inference has always considered as an important topic.
For instance: in image classification, 
novel architectures are regularly developed, 
Binarised Neural Networks~\cite{Courbariaux16, Rastegari16}, MobileNet~\cite{Howard2017mobile}, 
ShuffleNet~\cite{Zhang18-shufflenet}, 
EfficientNet~\cite{Tan2019eff}; 
and in object detection, single-stage detectors have drastically improved the efficiency of object detection,
{\em e.g.}, YOLO~\cite{Redmon16}, SSD~\cite{Liu16},
RetinaNet~\cite{Lin17}, CornerNet~\cite{Law18}, {\em etc}.
When processing videos, the efficiency becomes even more critical.
In the literature, approaches have been proposed to intelligently sample the video data, by exploiting the relatively cheap audio modality in videos~\cite{Korbar2019scsampler}, to
exploit variable mini-batch shapes for fast training~\cite{Wu20MultiGrid}, or to
directly train on compressed videos without decoding~\cite{Wu18}.\\[-0.8cm]

\paragraph{Efficient Transformers.}
Transformer model architectures have  gained immense interest due to their effectiveness on a range of tasks, including computer vision,
natural language processing, reinforcement learning, {\em etc.}
The self-attention mechanism is a critical design in Transformer 
that enables each input token to explicitly build relationships with others.
In order to improve the general efficiency, a rich set of ideas have been explored to reduce the complexity for computing the self-attention affinity matrix, for example, to develop recurrence in the Transformer~(Transformer-XL~\cite{Dai19xl}),
to reduce the computation consumption by only exploiting Transformer decoders with a controllable number of queries~(Perceiver~\cite{Jaegle21}),
to use learnable attention pooling~(Routing Transformer~\cite{Roy20Routing}, Sinkhorn Transformer~\cite{Tay2020sparse}, Reformer~\cite{kitaev2020reformer})
to rely on low rank matrix decomposition of the affinity matrix from self-attention~(Linformer~\cite{wang2020linformer}, MotionFormer~\cite{Patrick21MF}, Performer~\cite{Choromanski21performer}, Linear Transformer~\cite{Katharopoulos21Linear}), to exploit sparse representation~(Product Key Memory~\cite{lample2019key}) for the affinity matrix. In this paper, our proposal is orthogonal to the above mentioned techniques, specifically, motivated by the high redundancy in visual signals, we aim to train the Transformer on videos with only sparsely sampled visual patches. \\[-0.5cm]

%% file: sec/04-method.tex
\vspace{-3mm}
\section{Method}
\label{sec:method}
\vspace{-2mm}

\begin{figure}
    \centering
    \includegraphics[width=1.0\textwidth]{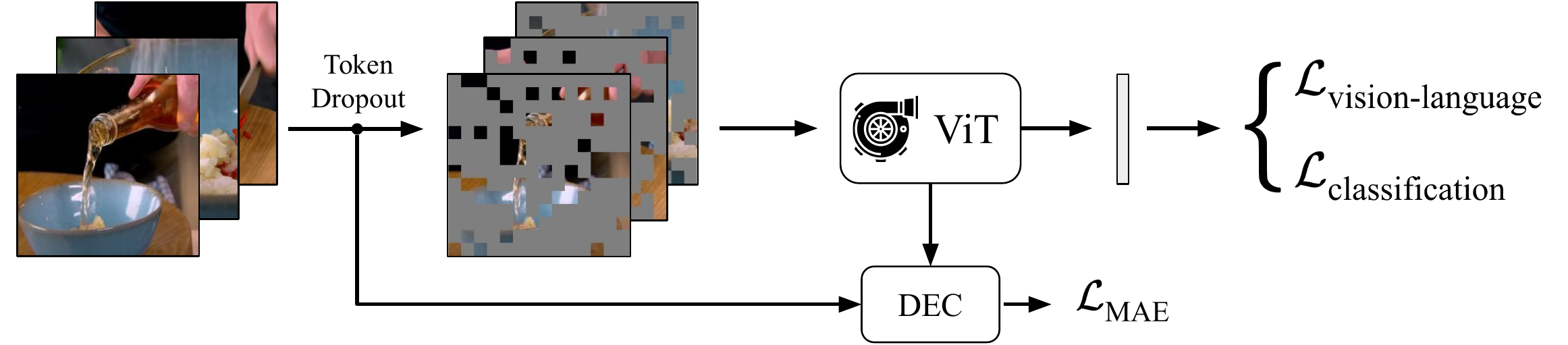}
    \vspace{-5mm}
    \caption{Illustration of Turbo training. 
    We  train a Video Transformer model by only using sparsely sampled visual tokens, for faster training and reduced computational cost. To achieve this goal, we exploit a multi-task training scheme, with one task being partial masked autoencoding,
    and the other being the standard supervised loss for the downstream task of interest, for example, contrastive learning for visual-language pre-training, or cross-entropy for action classification.}
    \label{fig:teaser}
    \vspace{-15pt}
\end{figure}

We start by presenting an overview of Vision Transformers in Section~\ref{sec:vit}, and discuss their pros and cons for video understanding tasks. In Section~\ref{sec:turbo}, we introduce the \textbf{Turbo} training, and describe its application for efficiently training the Vision Transformer on videos.

\vspace{-3mm}
\subsection{Vision Transformers for Videos} 
\label{sec:vit}
Generally speaking, given a video with $T$ frames, 
{\em i.e.}, $\mathcal{V} = \{I_1, I_2, \dots, I_T\}$, 
with $I_k \in \mathbb{R}^{H \times W \times 3}$,
as input to a Vision Transformer~(ViT), the video is first split into non-overlapping spatial-temporal patches, 
and projected with linear layers to get a sequence of token vectors. 
After appending a learnable `CLS' token~($v_\texttt{CLS}$), 
the resulting vector sequence is then processed with series of transformer encoder layers, 
{\em i.e.}, $\mathbf{z} = \Phi_{\textsc{ViT}}([v_\texttt{CLS}, v_1, ..., v_n ] + \texttt{PE})$, where $v_i$ denotes one of the $n = n_t \cdot n_h \cdot n_w$ spatio-temporal patches of size $t\times h\times w$, 
with $n_t = \lfloor T/t\rfloor, n_h = \lfloor H/h \rfloor, n_w = \lfloor W/w \rfloor$, and $\texttt{PE}$ refers to the spatio-temporal positional encodings.

With the self-attention operations along \emph{all} the spatial-temporal patches, the benefit of applying ViT on videos is prominent -- it can capture the long-term dependency in the videos.
However, the self-attention also incurs the quadratic complexity, 
{\em i.e.}, $\mathcal{O}((n_t\cdot n_h \cdot n_w)^2)$, 
for scenarios when the input video is long (more than a few seconds),
the memory consumption becomes infeasible.
As a consequence, the majority of ViT-based architectures are still limited to only processing short video clips.


\vspace{-2mm}
\subsection{Turbo Training}
\label{sec:turbo}

In this paper, we introduce a \textbf{Turbo training} regime, 
that aims to exploit the redundancy in video data,
and optimise a video Transformer for multi-task training with only sparsely sampled visual tokens. It requires optimizing for two objectives: 
(i) partial masked autoencoding, that acts as a proxy task to force the model to capture spatio-temporal information to a full extent, 
with only the sparsely sampled video patches; 
and (ii) the default objective for the considered downstream task, 
that may refer to cross-entropy or contrastive learning.

In the following, we start by describing the partial masked autoencoder,
and then detailing the use of Turbo training on various video understanding tasks, including: action classification, video-language training and long-video activity classification.
\\[-0.8cm]

\paragraph{Partial Masked Autoencoder (PMAE).}
We use a ViT as the visual encoder~($\Phi_{\text{ViT-enc}}(.)$),
that consists of a number of Transformer encoder blocks, 
operating on the input visual tokens.
Inspired by MAE~\cite{He2022MAE},
we also use a shallow decoder with much fewer Transformer blocks than the encoder~($\Phi_{\text{dec}}(.)$).
After preparing the spatial-temporal patches,
we define three operations for the embedded tokens: 
(1) masking the input visual tokens with a ratio $m$, 
which means only $N_i = n_t \cdot n_h \cdot n_w \cdot (1 - m)$ 
tokens are fed into the visual encoder~($\Phi_{\text{ViT-enc}}(.)$);
(2) reconstructing at a reconstruction ratio $r$,
that means $N_r = n_t \cdot n_h \cdot n_w \cdot r$ tokens are the targets of the reconstruction tasks;
(3) the unused $N_{\text{ignore}} =  n_t \cdot n_h \cdot n_w \cdot (m - r)$ tokens are ignored in the training iteration.
Note that a natural constraint $r \leq m$ holds, 
and the original MAE setting~\cite{He2022MAE} can be regarded as the special case when $r = m$. This training regime significantly cuts downs the memory consumption, for example, the complexity for self-attention in the encoder layer is only $\mathcal{O}(N_i^2)$, 
compared to operating on the full token sequence~($\mathcal{O}((n_t\cdot n_h\cdot n_w)^2)$), while the decoder layer can be quite lightweight, 
with the complexity of self-attention being $\mathcal{O}((N_i + N_r)^2)$.\\[-0.8cm]

\paragraph{Action Classification.}
Here, we apply the Turbo training for Transformers on the video action classification task. In addition to the above-mentioned partial MAE loss~($\mathcal{L}_{\texttt{PMAE}}$),
we adopt the standard cross-entropy loss on the short clips. 
Specifically, when given a labelled video sample $\{\mathcal{V},y\}$, 
{\em i.e.}, a short video clip with an action label, 
we take the `\texttt{CLS}' feature from the final layer of the visual encoder, then we pass the visual feature to a classifier, that is parametrised with a single MLP layer~(denoted as $g(.)$), and trained with cross entropy loss.
The overall objective for classification training is
\vspace{-1mm}
\begin{equation}
    \mathcal{L} = \lambda_{\texttt{CE}}\cdot \mathcal{L}_{\texttt{CE}}{(g(z_{\texttt{CLS}}), y)} 
+ \mathcal{L}_{\texttt{PMAE}}
\label{eq:loss_classification}
\end{equation}
\vspace{-2mm}
\noindent In practice we set the weight 
$\lambda_{\text{CE}} = 1 / \log{( \texttt{num\_classes} )}$ to balance two losses.

\vspace{-0.1cm}
\paragraph{Visual-language Pre-training.}
Recently, 
visual-language representation learning has attracted growing interest from the community, due to its convenience of data curation procedure and remarkable performance on ``zero-shot'' generalisation on various image classification tasks~\cite{Radford21,Jia21}.
However, the challenge for training the model lies in its requirement for a significant amount of compute and long training schedule. 
Turbo training can be applied to this scenario, 
to significantly speed up and reduce the memory cost.
In addition to the sparse MAE loss~($\mathcal{L}_{\texttt{PMAE}}$),
here we also adopt a standard InfoNCE loss~(noise contrastive learning~\cite{Oord18}) to learn the joint embedding for the visual and language streams.

Specifically, given a paired video-language sample $\{\mathcal{V}, y\}$, 
{\em e.g.}, a video clip and a sentence that describes the visual scene,
we use a Transformer-based language model $\Psi_{\text{text}}(\cdot)$ to encode the sentence, and take the language `\texttt{CLS}' token as the representation for the entire sentence:
\[ z_t = \Psi_{\text{text}}([w_{\texttt{CLS}}, w_1, \dots, w_n] + \texttt{PE})\]
\noindent where $w_i$ denotes word embedding after tokenisation, 
and $\texttt{PE}$ denotes the positional embedding for the language.
As for the visual stream, we use the same Transformer-based visual encoder~($\Phi_{\textsc{ViT-enc}}(\cdot)$), as introduced above, 
and take the final visual `\texttt{CLS}' feature as the visual feature.
The overall visual-language training loss is
\vspace{-1mm}
\begin{equation}
    \mathcal{L} = \lambda_{\texttt{NCE}} \cdot \mathcal{L}_{\texttt{NCE}} + \mathcal{L}_{\texttt{PMAE}}
\label{eq:loss_nce}
\vspace{-1mm}
\end{equation}

\noindent with $\mathcal{L}_{\texttt{NCE}}$ denoting a bi-directional~(visual-to-language \& language-to-visual) InfoNCE loss.
\[ \mathcal{L}_{\texttt{NCE}} = 
- \frac{1}{2} \bigg( \log \frac{\exp{(z_v \cdot z_t)}}{\sum_t{\exp{(z_v\cdot z_t)}}}
  + \log \frac{\exp{(z_v\cdot z_t)}}{\sum_{v}{\exp{(z_v\cdot z_t)}}}
  \bigg)\]
  
\noindent In practice we set the weight $\lambda_{\texttt{NCE}}=1 / \log{( \texttt{batch\_size} )}$ to balance two losses.

\vspace{-1mm}
\paragraph{Long-video Activity Classification.}

Turbo training enables to load longer video sequences for end-to-end training,
which was a challenging factor for long-video tasks.
By applying Turbo training, the method for
long-video activity classification is largely simplified and is 
similar to the method of short-video action classification.
Specifically,
given a labelled long video sample $\{\mathcal{V}, y\}$, 
we still optimise Eq.~\ref{eq:loss_classification} as the main objective, 
whilst applying a larger masking ratio on the input video $\mathcal{V}$
is necessary to reduce computational cost.

%% file: sec/05-experiment.tex
\vspace{-2mm}
\section{Experiments}
\label{sec:exp}
\vspace{-1mm}

In this section, we validate the effectiveness of Turbo training by experimenting on three different tasks:  video action classification,
video-language representation learning, and long-video activity classification.

\vspace{-2mm}
\subsection{Action Classification}\label{exp:act_classification}
\paragraph{Dataset.}
We conduct experiments on two datasets, 
\textbf{UCF101}~\cite{Soomro12}, containing 13k short video clips for 101 human actions, and \textbf{HMDB51}~\cite{Kuehne11}, containing 7k short video clips for 51 human actions. We report the Top1 classification accuracy on this task.\\[-0.8cm]

\paragraph{Implementation.}
In accordance with the experiment setting as used in~\cite{Tong2022VideoMAE},
for each short video clip we decode the video with 5 fps and randomly sample 16 video frames at $224\times 224$ resolution as input,
which approximately spans about 3.2 seconds in time.
The video frames are passed to the $\Phi_{\textsc{ViT}}(\cdot)$ with
a token dropout controlled by the mask ratio $m$, 
and we take the final-layer `CLS' token as the video feature,
which is further passed to a linear classifier and trained with cross-entropy~(CE) loss. Also, 
we keep a light-weight MAE objective controlled by a reconstruction ratio $r$, 
that is to reconstruct some randomly selected~($r\times 100\%$) space-time patches with a light weight ViT decoder.
The overall objective function is in equation~\ref{eq:loss_classification}.
We experiment with different combinations of the mask and reconstruction ratios $\{m, r\}$,
to study their effects on turbo training.
For more training details, we refer the reader to \app.

\vspace{-2mm}
\subsection{Video-Language Representation Learning}
\paragraph{Dataset.}
For the experiments on video-language pre-training,
we adopt the recent \textbf{HTM-AA} dataset,
where the video and language description have been temporally auto-aligned with self-supervision~\cite{Han22TAN}.
HTM-AA contains 250K videos sourced from the HowTo100M~\cite{Miech19} dataset, 
with over 3.3M clip-sentence pairs. 
For evaluation of the learnt video-language representation, 
we benchmark on a temporal action alignment task on the \textbf{HTM-Align} dataset~\cite{Han22TAN}. Specifically, HTM-Align contains 80 long videos 
from YouTube with a total of 5K ASR sentences, 
and each visually alignable sentence has a \textbf{manually} labelled temporal segment. 
Note that, we only use this HTM-Align dataset for evaluation purposes, 
and all training is done on the automatically curated HTM-AA dataset, 
thus it can be seen as a `zero-shot' evaluation on the HTM-Align.\\[-0.8cm]

\paragraph{Implementation.}
In accordance with~\cite{Lin22DistSup},
we choose MPNet~\cite{song2020mpnet} as the language encoder.
MPNet takes as input a sentence and outputs a textual feature vector.
On the visual side, we use $\Phi_{\textsc{ViT}}(\cdot)$ and take the final-layer `CLS' token as the video feature.
To demonstrate the effectiveness of Turbo training for the Vision Transformer, 
we fix the weights of the language model, and only finetune the weights of the ViT architecture, initialized from VideoMAE~\cite{Tong2022VideoMAE}~(pretrained on Kinetics-400~\cite{Kay17}).
As in Sect.~\ref{exp:act_classification}, 
the visual encoder takes as input a 16-frame short video clip extracted with 5 fps, and the language encoder takes the corresponding sentence associated with the video clip.
Overall, we take a batch size of at most 32 clip-sentence pairs~(depending on different masking ratios) for video-language training, with an InfoNCE loss.
We refer the reader to \app\ for more training and evaluation details.

\vspace{-2mm}
\subsection{Long-video Activity Classification}
\paragraph{Dataset.}
To demonstrate the effectiveness of exploiting Turbo training for end-to-end Transformer training on long video sequences,
here we evaluate long-video activity classification on the Breakfast~\cite{Kuehne12} and COIN~\cite{Tang2019coin} datasets.
In detail, \textbf{Breakfast} contains 1712 long videos for 10 cooking activities including `making coffee' and `frying egg'. 
\textbf{COIN} contains 11k long videos covering 180 general activities including `car polishing' and `assembling sofa'. 
The videos in these two datasets are often minute-long and untrimmed, 
containing procedural actions to complete the particular activity. \\[-0.8cm]

\paragraph{Implementation.}
In our experiment, we load $n=\{16, 32, 64\}$ frames as the video input, 
and train the Transformer end-to-end.
Note that, such an end-to-end setting was not easily achievable in 
previous works due to the large memory footprint.
In detail, at the training time, we randomly choose the start and end timestamp among the first and last $20\%$ duration of the video, 
then uniformly take $n$ frames in between.
The rest of the architectural design is similar to that of Sect.~\ref{exp:act_classification}.
At inference time, we repeat sampling $n$ frames for 10 times from each video,
and average the prediction probability,
resembling the multi-crop test that has been widely used in existing work~\cite{Simonyan14b}.
By default, we use a batch size of 16 videos and train the Transformer for 100 epochs on Breakfast and 50 epochs on COIN. 

%% file: sec/06-result.tex
\vspace{-3mm}
\section{Results}
\vspace{-2mm}

Here, we start by showing the efficiency improvement from Turbo training on action classification~(Sect.~\ref{sec:ablation}),
then compare with existing works~(Sect.~\ref{sec:result_short}).
Next, we report the results for video-language representation learning (Sect.~\ref{sec:result_pt}),
followed by long-video activity classification (Sect.~\ref{sec:result_long}).

\vspace{-5pt}
\subsection{Turbo Training for Action Classification}
\subsubsection{Ablation Study: the speed and performance trade-off}
\label{sec:ablation}

In action classification,
we investigate two of the hyperparameters of the PMAE, 
namely the mask ratio $m$ and reconstruction ratio $r$.
For the training, 
we use ViT-B initialized with the VideoMAE weights,
and then finetune the network on UCF101 for action classification.\\[-0.8cm]

\input{table/speed_vs_acc}
\input{table/ratios}  

\paragraph{Both high mask ratio and low reconstruction ratio can reduce the computation.}
In Table~\ref{table:speed-acc}~(left),
we compute the GFLOPs for the Transformer model during Turbo training.
Note that GFLOPs contains both the classifier head for action classification 
and the decoder for patch reconstruction.
It is evident that increasing the masking ratio can reduce the GFLOPs,
{\em e.g.}, by applying a 50\% masking ratio for action classification,
the GFLOPs can be reduced by half at a cost of only 1\% Top1 classification accuracy. Additionally, we find reducing the reconstruction ratio can also reduce
the GFLOPs while incurring minimum effect to the final classification accuracy (91.8 to 91.6, and 89.7 to 89.6).
In the following experiments,
we use settings with $\{m,r\}=\{0.75,0.25\}$ and $\{0.5,0.5\}$, that demonstrate a balance between the performance and the compute operations. Figure~\ref{fig:mem} and~\ref{fig:finetune} show the
reduced memory footprints and the significant speed-up during training.\\[-0.8cm]

\paragraph{Generalisation to smaller masking ratio during inference.}
Here, we take the models trained with $\{m,r\} = \{0.9,0.1\}$ and $\{0.75,0.25\}$, and test it with different masking ratio at inference time.
In Table~\ref{table:speed-acc}~(right),
we show that the model trained with high-rate token dropout can be used with a lower mask ratio at inference time.
For example, inference with full patches ($m=0$)
achieves the best performance regardless of the training masking ratio.
In the following experiments, we thus use $m=0$ at inference by default.

\vspace{-4mm}
\subsubsection{Comparison to state-of-the-art}
\label{sec:result_short}
In Table~\ref{table:classification}, we compare to the existing approaches, 
which have been pre-trained on video or multimodal data in a self-supervised manner, and then end-to-end finetuned on the downstream tasks.
\textbf{Note that}, the comparison here is by no means fair, 
due to the differences in architecture, training data, and resolutions.
However, the key message in this comparison is that, 
even with only 50\% of the visual tokens for finetuning, 
the model can still achieve competitive performance compared to all other models, 
while only needing half the training time as the original finetuning with all visual tokens. Additionally, with further self-supervised Turbo training on video-language data~(in Sect.~\ref{sec:result_pt}), 
the performance can be further boosted.

\input{table/ucf_hmdb}

\subsection{Turbo Training for Video-Language Representation Learning}
\label{sec:result_pt}
In this section, we experiment with video-language turbo training on the HTM-AA dataset.
In detail, we consider three settings: 
no masking, $\{m,r\} = \{0.5,0.5\}$, and $\{0.75,0.25\}$.
For each setting, we use the same implementation 
and the maximum number of video sequences a single NVidia A40 GPU can handle. We measure the quality of the video-language representation by monitoring the temporal action alignment performance on the HTM-Align dataset~\cite{Han22TAN},
{\em i.e.}, training wall-clock time vs.~the temporal alignment performance.
In detail, we extract the per-second visual features for the long videos in HTM-Align,  then use the text embeddings to retrieve the corresponding visual moment in the time axis, recall is 1.0 if the predicted timestamp fall into the ground-truth temporal segment. We report Recall@1 (R@1) only on alignable sentences as in~\cite{Han22TAN}.

\vspace{1mm}
\noindent \textbf{Evaluation Details.}
We follow the evaluation method of~\cite{Han22TAN} that was applied on the CLIP and MIL-NCE features.  For each minute-long video,
there are two steps to evaluate the backbone feature quality on temporal alignment:
(1) use the visual backbone to extract video features for the whole video at 1 feature-per-second, resulting in a sequence of vectors;
(2) use the given sentence embedding to compute the similarity score with the sequence of visual features, and determine the most corresponding timestamps.
In this paper, we adopt the same evaluation setting, specifically, 
we extract the video frames at 16 fps and pass every 
16 frames into our ViT to compute a single feature vector,
resulting in a sequence of visual features at 1 feature-per-second.
Therefore, in Table~\ref{table:vl} our temporal receptive field (R.F.)
is 1 second because there is no long-range temporal modelling in this evaluation.

\vspace{1mm}
\noindent \textbf{Discussion.}
As shown in Figure~\ref{fig:vl}, 
it is clear that a higher mask ratio significantly increases the training efficiency.
For example, after being trained for 18h, the 0.75 mask ratio achieves 28\% R@1, 
whereas the baseline setup only gets 11\% R@1.
Additionally, to get the same performance,
the $m=0.75$ setting gives a 3.5$\times$ speed-up comparing with $m=0.5$ 
(reaching 27\% R@1 requires 11.3 hours and 40 hours respectively).
For the baseline setting without Turbo training,
it is not feasible to achieve a comparable performance under a single GPU training budget, 
showing that Turbo training indeed leads to improved performance and new abilities for existing Transformer models to finetune on video tasks in an end-to-end manner.

\begin{table}[!htb]
    \centering
    \hspace{-1.0cm}
        \centering
        \small
        \includegraphics[width=0.6\textwidth]{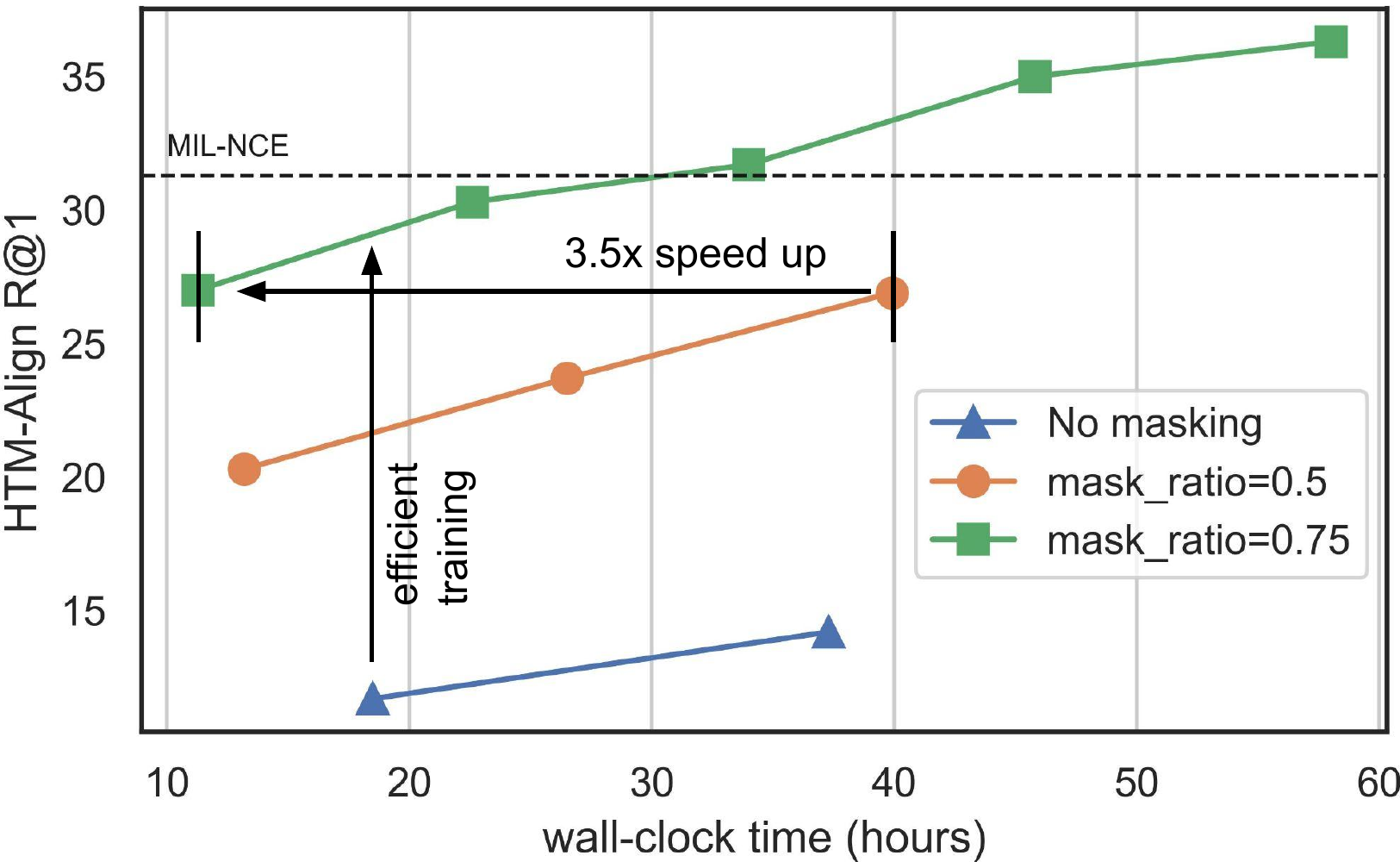}
        \caption{Progress of vision-language pretraining 
        monitored by R@1 on HTM-Align dataset.}
    \label{fig:vl}
\end{table}

\begin{table}[!htb]
        \centering
        \small
        \begin{tabu}{c|c|c|c|c}
        \hline
        Method & Arch. & Pretrain Data & Temporal R.F. & R@1 \\ \hline
        CLIP~\cite{Radford21} & ViT-B-32 & YFCC & 1s & 16.8 \\
        CLIP~\cite{Radford21} & ViT-B-16 & YFCC & 1s & 22.0 \\
        MIL-NCE~\cite{Miech20} & S3D-G & HTM & 1s & 31.3 \\
        NCE with TAN$^{\dag}$~\cite{Han22TAN} & S3D-G & HTM-AA  & 1s & {33.6}\\
        Ours (m=0.75, r=0.25) & ViT-B-16 & HTM-AA & 1s & \textbf{36.7} \\ \hline
        \rowfont{\color{gray}}
        TAN$^{*}$~\cite{Han22TAN} & S3D-G \& TFM  & HTM & 64s & 49.4 \\ \bottomrule
        \end{tabu}
     \caption{
    Zero-shot text-visual alignment results on the HTM-Align dataset. 
    We adopt the same setting as~\cite{Han22TAN} and report Recall@1 as the metric.
    $*$: the result from TAN is not directly comparable
    as their model contains transformers (TFM) on the  S3D-G features and has a much longer temporal context (64s) for the temporal alignment task.
    $\dag$: we train the NCE loss on HTM-AA for the same duration (approx.\ 60h) for a fair comparison. }
    \label{table:vl}
    \vspace{3mm}
\end{table}

\vspace{-5mm}
In Table~\ref{table:vl},
our video-language representation gets a higher R@1 (36.7) than previous strong methods, including CLIP~(22.0) and MIL-NCE~(31.3), 
and the S3D model after end-to-end finetuning on the HTM-AA dataset~(33.6),
showing the superior quality of the joint video-language embedding learnt from Turbo training. 
Note that, the best alignment results~(49.4) from TAN~\cite{Han22TAN}
is not directly comparable with our method,
as their model builds on pre-extracted S3D base features and 
learns long-range temporal Transformers that scan the 
surrounding 64-second temporal context for the temporal alignment task.
Here, we aim to evaluate the quality of \emph{base} features, 
which is orthogonal to the long-term modelling in TAN~\cite{Han22TAN}.

\vspace{-2mm}
\subsection{Turbo Training for Long-video Activity Classification}
\label{sec:result_long}

Turbo training enables the end-to-end training of Transformers on long-video tasks, that used to be a critical bottleneck.
We compare three settings: 
\textbf{(1)} $F_{16}$ denotes 16-frame input with $\{m,r\} = \{0.5,0.5\}$,
\textbf{(2)} $F_{32}$ denotes 32-frame input with $\{m,r\} = \{0.75,0.25\}$, and
\textbf{(3)} $F_{64}$ denotes 64-frame input with $\{m,r\} = \{0.875,0.125\}$.
The number of frames and masking ratios are chosen such that
each of the setting has the same number of \emph{visible} 
patches to the encoder but with different temporal extent. 
For frame sampling, 
we first sample the starting and ending positions from the beginning and the last 20\% of the video, then we uniformly sample frames within this segment.\\[-0.8cm]

\paragraph{Effect of loading longer sequences.}
In Table~\ref{table:breakfast}, 
increasing the frame numbers from 16 to 32 (row B vs.\ C) shows the longer temporal extent is beneficial for activity classification
(e.g. 91.3 vs.\ 88.2 on Breakfast),
we conjecture a denser temporal sampling has a higher chance to 
sample the characteristic moment in the long video.
However, further increasing the frame numbers to 64
does not help the activity classification task.
It might be because the high token masking ratio ($m=0.875$)
could frequently miss the key objects or actions in the video frame.

\input{table/breakfast_coin}

\paragraph{Comparison with other works.}
When learning from 32 frames on the Breakfast dataset end-to-end,
our simple one-stage method outperforms previous methods (91.3 vs.\ 89.9) 
which typically require a two-stage setting:
extracting short-clip features, then training temporal models on these pre-extracted features.
On the COIN dataset, 
our one-stage method achieves comparable results (87.5 vs.\ 88.9).
We conjecture a longer temporal context with smaller masking ratio
could further improve the performance on uncurrated natural videos like COIN, 
which inevitably demands higher hardware requirements.
Comparing row C and row A further shows the effectiveness of
our vision-language training from Sect.~\ref{sec:result_pt}.

%% file: table/speed_vs_acc.tex

\begin{table}[t]
    \centering
    \vspace{-0mm}
    \begin{minipage}{0.5\textwidth}
        \centering
        \footnotesize
        \begin{tabular}{lll|l}
            \hline
            Mask\%  & Recon\% & GFLOPs & Acc (\%) \\  \hline
            0   & - (w/o \texttt{MAE})   & 180.6     & 94.8$^*$    \\\hline
            50  & 50 (\texttt{MAE})  & 99.3   & 93.7    \\ \hline
            75  & 75 (\texttt{MAE})  & 57.6   & 91.8    \\ 
            75  & 25 (\texttt{PMAE}) & 45.9   & 91.6    \\ \hline
            90  & 90 (\texttt{MAE})  & 35.2   & 89.7    \\ 
            90  & 10 (\texttt{PMAE}) & 18.3   & 89.6    \\ \hline
        \end{tabular}
    \end{minipage}
    \begin{minipage}{0.45\textwidth}
        \centering
        \footnotesize
        \begin{tabular}{c|c|c}
        \hline
           Train ($m$\%,$r$\%) & Inference $m$\% & Acc (\%) \\ \hline
           90, 10  & 90 & 89.6 \\
           90, 10  & 50 & 90.1 \\
           90, 10  & 0 &  \textbf{90.3} \\ \hline
           75, 25  & 75 & 91.6 \\
           75, 25  & 50 & 91.9 \\
           75, 25  & 0  & \textbf{92.4} \\
           \hline
        \end{tabular}
    \end{minipage}
    \caption{
    \textbf{Left}: Speed and performance trade-off on the UCF101 classification task. 
    *: our implementation of~\cite{Tong2022VideoMAE} on UCF101.
    \textbf{Right}: Generalisation to different mask ratios at the inference time.
    }
    \vspace{5mm}
    \label{table:speed-acc}
\end{table}

%% file: table/ratios.tex
\begin{figure}[t]
    \centering
    \vspace{-3.5mm}
    \begin{minipage}{0.44\textwidth}
        \centering
        \includegraphics[height=0.72\textwidth]{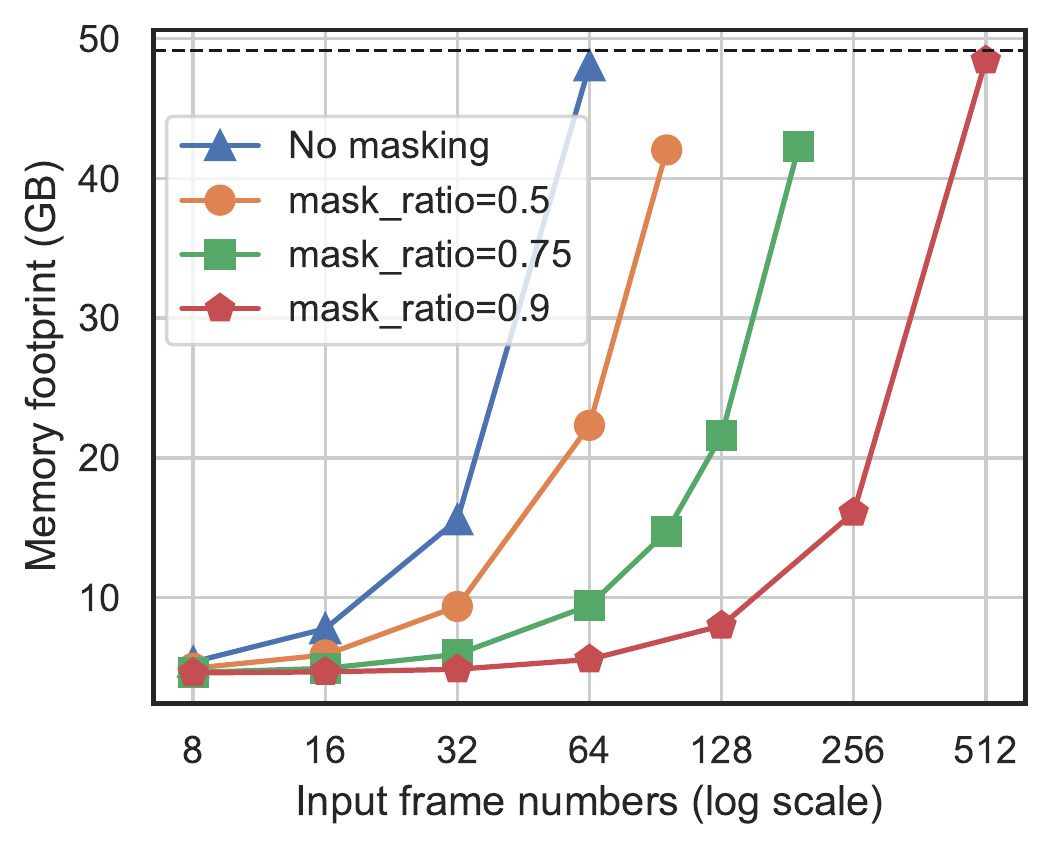}
        \vspace{-1mm}
        \caption{Memory footprint of training one video with different input frames.}
        \label{fig:mem}
    \end{minipage}
    \begin{minipage}{0.54\textwidth}
        \centering
        \includegraphics[height=0.58\textwidth]{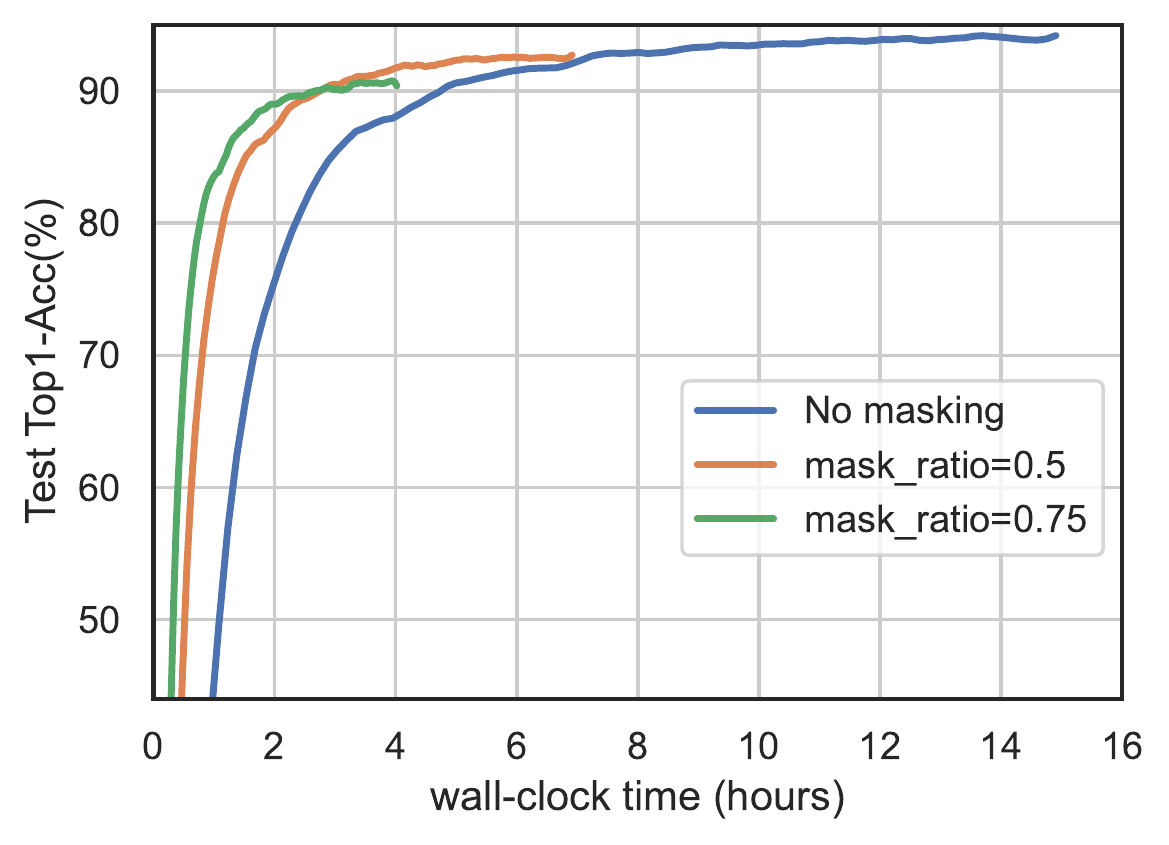}
        \vspace{-1mm}
        \caption{
        Training progress on UCF101 with a batch size of 16 for 100 epochs.}
        \label{fig:finetune}
    \end{minipage}
    \vspace{-6.5mm}
\end{figure}

%% file: table/ucf_hmdb.tex
\begin{table}[t]
\centering
\footnotesize
\begin{tabu}{lllll|ll}
\hline
Finetune Method & Pretrain Method   & Backbone  & Pretrain Dataset & Modality & UCF101 & HMDB51 \\ \hline
Classic & CoCLR~\cite{Han20coclr} & S3D       & K400        & V    & 87.9          & 54.6 \\
Classic & CVRL~\cite{Qian20}      & Slow-R152 & K600        & V    & 93.6          & 69.4 \\
Classic & $\rho$-BYOL~\cite{Feichtenhofer2021large} & Slow-R50  & K400        & V    & 94.2          & 72.1 \\
Classic$\dagger$ & VideoMAE~\cite{Tong2022VideoMAE} & ViT-B     & K400        & V    & {94.7}          & - \\ 
\rowfont{\color{gray}}
Classic & VideoMAE~\cite{Tong2022VideoMAE} & ViT-B     & K400+UCF101        & V    & {96.1} & 73.3 \\ 
\hline
Classic & MIL-NCE~\cite{Miech20}  & S3D-G       & HTM         & V+T  & 91.3          & 61.0 \\
Classic & TAN~\cite{Han22TAN}     & S3D-G       & HTM-AA   & V+T  & 92.0          & - \\
Classic & XDC~\cite{Alwassel20}   & R(2+1)D-18  & IG65M       & V+A  & 94.2          & 67.1 \\
Classic & GDT~\cite{Patrick2020gdt}& R(2+1)D-18  & IG65M       & V+A  & 95.2          & 72.8 \\ \hline
\textbf{Turbo}$_{m=0.5,r=0.5}$ & VideoMAE~\cite{Tong2022VideoMAE}  & ViT-B & K400  & V    & 94.4          & 71.8 \\
\textbf{Turbo}$_{m=0.5,r=0.5}$ & NCE  & ViT-B & HTM-AA      & V+T  & \textbf{95.7}          & \textbf{73.6} \\
\hline
\end{tabu}
\caption{Comparing with recent self-supervised methods by finetuning on UCF101 and HMDB51 action classification.
We apply Turbo training for finetuning, `Classic' means traditional finetuning method without any token dropout or not applicable to token dropout.
$\dagger$: our reproduction of finetuning the open-sourced model 
from~\cite{Tong2022VideoMAE} which was trained on K400 only.
{\color{gray}The grey row}: The best result on UCF101 from~\cite{Tong2022VideoMAE} is pretrained on UCF101 for an additional 3200 epochs, therefore it is different to their open-sourced model~(only trained on K400). We use $\{m,r\}=\{0.5,0.5\}$ for our Turbo training settings.}
\vspace{-3mm}
\label{table:classification}
\end{table}



%% file: table/breakfast_coin.tex
\begin{table}[h]
\centering
\footnotesize
\begin{tabular}{llll|ll}
\hline
Temporal Method      & Clip Model       & Pretrain Supervision            & Pretrain Data & BF & COIN \\ \hline
Timeception~\cite{Hussein19timeception} & 3D-ResNet           & [\cmark] action labels & K400 & 71.3 & -    \\
VideoGraph~\cite{Hussein19graph} & I3D                 & [\cmark] action labels & K400 & 69.5 & -   \\
GHRM~\cite{Zhou21GHRM}        & I3D                 & [\cmark] action labels & K400 & 75.5 & -   \\
Transformer~\cite{Lin22DistSup} & TimeSformer   & [\cmark] action labels & K400 & 81.1 & 83.5  \\ 
Transformer~\cite{Lin22DistSup} & TimeSformer   & [\xmark] k-mean on ASR & HTM & 81.4 & 85.3   \\
Transformer~\cite{Lin22DistSup} & TimeSformer   & [\xmark] Dist. Sup & HTM+WikiHow & 89.9 & \textbf{88.9}   \\ \hline
{}[A] \textbf{Turbo} $F_{32}$      & N/A (end-to-end)   & [\xmark] VideoMAE & K400 & 86.8 & 82.3       \\ 
{}[B] \textbf{Turbo} $F_{16}$      & N/A (end-to-end)   & [\xmark] NCE & HTM-AA  & 88.2 & 81.2           \\ 
{}[C] \textbf{Turbo} $F_{32}$      & N/A (end-to-end)   & [\xmark] NCE & HTM-AA  & \textbf{91.3} & 87.5       \\
{}[D] \textbf{Turbo} $F_{64}$      & N/A (end-to-end)   & [\xmark] NCE & HTM-AA  & 87.3 & 86.8  \\
\hline
\end{tabular}
\caption{Procedural activity classification on the Breakfast~(BF) and COIN dataset.
[\cmark] denotes supervised training and 
[\xmark] denotes training without manual labelling.
}
\vspace{-7mm}
\label{table:breakfast}
\end{table}

%% file: sec/07-conclusion.tex
\vspace{-3mm}
\section{Conclusion}
\vspace{-3mm}

We propose a simple and versatile Turbo training paradigm
for Vision Transformers.
We show that it is applicable to multiple video tasks 
including action classification,
video-language training, and long-video activity classification,
and can achieve superior performance, 
while significantly reducing the required computation and memory cost.
Turbo training has demonstrated the ability to lower the resource requirements and enable end-to-end Transformer training for long videos.

\subsection*{Acknowledgements}
This research is funded by the EPSRC Programme Grant 
VisualAI EP/T028572/1, 
a Royal Society Research Professorship RP\textbackslash R1\textbackslash191132, 
and a Google-DeepMind Scholarship.
We thank Charig Yang and Chuhan Zhang for proof-reading and discussion.

%% file: sec/08-supp.tex
\vspace{-4mm}

We provide the implementation details in Sect.~\ref{sec:imp}.
For the code and models of this paper,
please refer to our project page: {\footnotesize\url{https://www.robots.ox.ac.uk/~vgg/research/turbo/}}.

\vspace{-0.3cm}
\section{Implementation Details}
\label{sec:imp}

\paragraph{Architectural Details.}
In our implementation, 
we adopt the standard ViT-B architectures as~\cite{He2022MAE,Tong2022VideoMAE}.
Specifically, the encoder is a 12-layer transformer with 768 feature dimension
and the light-weight decoder is a 8-layer transformer with 512 feature dimension.
The input spatial-temporal patch has a size of $t\times h\times w = 2\times 16\times 16$.
We use sinusoidal positional embeddings~\cite{Vaswani17}.
For both the action classification and long-video activity classification tasks,
we pass the encoder's final-layer `CLS' token into a linear layer for classification.
For learning video-language representation,
we project both the video feature and language feature with a 2-layer MLP, then compute the InfoNCE loss $\mathcal{L}_{\texttt{NCE}}$ as introduced in the main paper Page 5.

\input{table/supp_impl}

\paragraph{Training Details.}
The details of training action classification, 
video-language training and long-video activity classification tasks are listed in Table~\ref{tab:supp_implementation}.
Note that, for action classification and long-video activity classification tasks,
we use the same data augmentation as in~\cite{He2022MAE,Tong2022VideoMAE};
for video-language training, we only use basic cropping augmentation 
due to the adequate amount of training data from the HTM-AA~\cite{Han22TAN} dataset
(3.3M clip-sentence pairs).

%% file: table/supp_impl.tex
\begin{table}[h]
    \centering
    \scriptsize
    \begin{tabu}{c|c|c|c}
        \hline
           Config & Act. Classification & V-L Training & Long-video Activity Classification \\ \hline
        ViT-B encoder depth & 12 layers  & 12 layers & 12 layers \\
        ViT-B encoder dimension & 768 & 768 & 768 \\
        decoder depth & 8 layers  & 8 layers & 8 layers \\
        decoder dimension & 512 & 512 & 512 \\ \hline
        optimizer & AdamW~\cite{Loshchilov19adamw} & AdamW & AdamW \\
        base learning rate & 1e-3 & 1e-4 & 3e-4 \\
        weight decay & 0.05 & 0.05 & 0.05 \\
        learning rate schedule & cosine-decay~\cite{loshchilov2017sgdr}  & cosine-decay & cosine-decay \\
        warm-up epochs & 10 & 0.5 & 10(BF), 5(COIN)  \\
        training epochs & 100 & 5 & 100(BF), 50(COIN) \\
        repeated sampling~\cite{Hoffer20Repeat,Feichtenhofer2022VideoMAE} & 1 & 4 & 4 \\
        augmentation & RandAug(9,0.5)~\cite{Cubuk20randaug} & MultiScaleCrop & RandAug(9,0.5) \\
        label smoothing~\cite{Szegedy16} & 0.1 & - & 0.1 \\
        mixup~\cite{zhang2018mixup} & 0.8 & - & 0.8 \\
        cutmix~\cite{Yun2019Cutmix} & 1.0 & - & 1.0 \\
        drop path~\cite{Huang16droppath} & 0.1 & 0.0 & 0.1 \\
        \hline
    \end{tabu}
    \caption{
    Implementation details of action classification, video-language training
    and long-video activity classification tasks.}
    \label{tab:supp_implementation}
    \vspace{-0.8cm}
\end{table}

%% file: main.bbl
\begin{thebibliography}{62}
\providecommand{\natexlab}[1]{#1}
\providecommand{\url}[1]{\texttt{#1}}
\expandafter\ifx\csname urlstyle\endcsname\relax
  \providecommand{\doi}[1]{doi: #1}\else
  \providecommand{\doi}{doi: \begingroup \urlstyle{rm}\Url}\fi

\bibitem[Alwassel et~al.(2020)Alwassel, Mahajan, Torresani, Ghanem, and
  Tran]{Alwassel20}
Humam Alwassel, Dhruv Mahajan, Lorenzo Torresani, Bernard Ghanem, and Du~Tran.
\newblock Self-supervised learning by cross-modal audio-video clustering.
\newblock In \emph{NeurIPS}, 2020.

\bibitem[Bao et~al.(2022)Bao, Dong, Piao, and Wei]{Bao22BEiT}
Hangbo Bao, Li~Dong, Songhao Piao, and Furu Wei.
\newblock {BEiT}: Bert pre-training of image transformers.
\newblock In \emph{Proc. ICLR}, 2022.

\bibitem[Choromanski et~al.(2021)Choromanski, Likhosherstov, Dohan, Song, Gane,
  Sarl{\'{o}}s, Hawkins, Davis, Mohiuddin, Kaiser, Belanger, Colwell, and
  Weller]{Choromanski21performer}
Krzysztof Choromanski, Valerii Likhosherstov, David Dohan, Xingyou Song,
  Andreea Gane, Tam{\'{a}}s Sarl{\'{o}}s, Peter Hawkins, Jared Davis, Afroz
  Mohiuddin, Lukasz Kaiser, David Belanger, Lucy~J. Colwell, and Adrian Weller.
\newblock Rethinking attention with performers.
\newblock In \emph{Proc. ICLR}, 2021.

\bibitem[Courbariaux et~al.(2016)Courbariaux, Hubara, Soudry, El-Yaniv, and
  Bengio]{Courbariaux16}
Matthieu Courbariaux, Itay Hubara, Daniel Soudry, Ran El-Yaniv, and Yoshua
  Bengio.
\newblock Binarized neural networks: Training deep neural networks with weights
  and activations constrained to +1 or -1.
\newblock \emph{arXiv preprint arXiv:1602.02830}, 2016.

\bibitem[Cubuk et~al.(2020)Cubuk, Zoph, Shlens, and Le]{Cubuk20randaug}
Ekin~Dogus Cubuk, Barret Zoph, Jon Shlens, and Quoc Le.
\newblock Randaugment: Practical automated data augmentation with a reduced
  search space.
\newblock In \emph{NeurIPS}, 2020.

\bibitem[Dai et~al.(2019)Dai, Yang, Yang, Carbonell, Le, and
  Salakhutdinov]{Dai19xl}
Zihang Dai, Zhilin Yang, Yiming Yang, Jaime Carbonell, Quoc~V. Le, and Ruslan
  Salakhutdinov.
\newblock Transformer-{XL}: Attentive language models beyond a fixed-length
  context.
\newblock 2019.

\bibitem[Devlin et~al.(2018)Devlin, Chang, Lee, and Toutanova]{Devlin18}
Jacob Devlin, Ming{-}Wei Chang, Kenton Lee, and Kristina Toutanova.
\newblock Bert: Pre-training of deep bidirectional transformers for language
  understanding.
\newblock \emph{arXiv preprint arXiv:1810.04805}, 2018.

\bibitem[Dosovitskiy et~al.(2020)Dosovitskiy, Beyer, Kolesnikov, Weissenborn,
  Zhai, Unterthiner, Dehghani, Minderer, Heigold, Gelly, Uszkoreit, and
  Houlsby]{Dosovitskiy20}
Alexey Dosovitskiy, Lucas Beyer, Alexander Kolesnikov, Dirk Weissenborn,
  Xiaohua Zhai, Thomas Unterthiner, Mostafa Dehghani, Matthias Minderer, Georg
  Heigold, Sylvain Gelly, Jakob Uszkoreit, and Neil Houlsby.
\newblock An image is worth 16x16 words: Transformers for image recognition at
  scale.
\newblock In \emph{Proc. ICLR}, 2020.

\bibitem[Feichtenhofer et~al.(2021)Feichtenhofer, Fan, Xiong, Girshick, and
  He]{Feichtenhofer2021large}
Christoph Feichtenhofer, Haoqi Fan, Bo~Xiong, Ross Girshick, and Kaiming He.
\newblock A large-scale study on unsupervised spatiotemporal representation
  learning.
\newblock In \emph{Proc. CVPR}, 2021.

\bibitem[Feichtenhofer et~al.(2022)Feichtenhofer, Fan, Li, and
  He]{Feichtenhofer2022VideoMAE}
Christoph Feichtenhofer, Haoqi Fan, Yanghao Li, and Kaiming He.
\newblock Masked autoencoders as spatiotemporal learners.
\newblock \emph{arXiv preprint arXiv:2205.09113}, 2022.

\bibitem[Gupta et~al.(2022)Gupta, Tian, Zhang, Wu, Martín-Martín, and
  Fei-Fei]{Gupta22maskvit}
Agrim Gupta, Stephen Tian, Yunzhi Zhang, Jiajun Wu, Roberto Martín-Martín,
  and Li~Fei-Fei.
\newblock {MaskViT}: Masked visual pre-training for video prediction.
\newblock \emph{arXiv preprint arXiv:2206.11894}, 2022.

\bibitem[Han et~al.(2020)Han, Xie, and Zisserman]{Han20coclr}
Tengda Han, Weidi Xie, and Andrew Zisserman.
\newblock Self-supervised co-training for video representation learning.
\newblock In \emph{NeurIPS}, 2020.

\bibitem[Han et~al.(2022)Han, Xie, and Zisserman]{Han22TAN}
Tengda Han, Weidi Xie, and Andrew Zisserman.
\newblock Temporal alignment networks for long-term video.
\newblock In \emph{Proc. CVPR}, 2022.

\bibitem[He et~al.(2022)He, Chen, Xie, Li, Doll{\'a}r, and Girshick]{He2022MAE}
Kaiming He, Xinlei Chen, Saining Xie, Yanghao Li, Piotr Doll{\'a}r, and Ross
  Girshick.
\newblock Masked autoencoders are scalable vision learners.
\newblock In \emph{Proc. CVPR}, 2022.

\bibitem[Hoffer et~al.(2020)Hoffer, Ben{-}Nun, Hubara, Giladi, Hoefler, and
  Soudry]{Hoffer20Repeat}
Elad Hoffer, Tal Ben{-}Nun, Itay Hubara, Niv Giladi, Torsten Hoefler, and
  Daniel Soudry.
\newblock Augment your batch: better training with larger batches.
\newblock In \emph{Proc. CVPR}, 2020.

\bibitem[Howard et~al.(2017)Howard, Zhu, Chen, Kalenichenko, Wang, Weyand,
  Andreetto, and Adam]{Howard2017mobile}
Andrew~G. Howard, Menglong Zhu, Bo~Chen, Dmitry Kalenichenko, Weijun Wang,
  Tobias Weyand, Marco Andreetto, and Hartwig Adam.
\newblock {MobileNets}: Efficient convolutional neural networks for mobile
  vision applications.
\newblock \emph{arXiv preprint arXiv:1704.04861}, 2017.

\bibitem[Huang et~al.(2016)Huang, Sun, Liu, Sedra, and
  Weinberger]{Huang16droppath}
Gao Huang, Yu~Sun, Zhuang Liu, Daniel Sedra, and Kilian~Q. Weinberger.
\newblock Deep networks with stochastic depth.
\newblock In \emph{Proc. ECCV}, 2016.

\bibitem[Hussein et~al.(2019{\natexlab{a}})Hussein, Gavves, and
  Smeulders]{Hussein19graph}
Noureldien Hussein, Efstratios Gavves, and Arnold~W.M. Smeulders.
\newblock {VideoGraph}: Recognizing minutes-long human activities in videos.
\newblock In \emph{ICCV-Workshop}, 2019{\natexlab{a}}.

\bibitem[Hussein et~al.(2019{\natexlab{b}})Hussein, Gavves, and
  Smeulders]{Hussein19timeception}
Noureldien Hussein, Efstratios Gavves, and Arnold~W.M. Smeulders.
\newblock Timeception for complex action recognition.
\newblock In \emph{Proc. CVPR}, 2019{\natexlab{b}}.

\bibitem[Ilya~Loshchilov(2019)]{Loshchilov19adamw}
Frank~Hutter Ilya~Loshchilov.
\newblock Decoupled weight decay regularization.
\newblock In \emph{Proc. ICLR}, 2019.

\bibitem[Jaegle et~al.(2021)Jaegle, Gimeno, Brock, Zisserman, Vinyals, and
  Carreira]{Jaegle21}
Andrew Jaegle, Felix Gimeno, Andrew Brock, Andrew Zisserman, Oriol Vinyals, and
  Jo{\~{a}}o Carreira.
\newblock Perceiver: General perception with iterative attention.
\newblock In \emph{Proc. ICML}, 2021.

\bibitem[Jia et~al.(2021)Jia, Yang, Xia, Chen, Parekh, Pham, Le, Sung, Li, and
  Duerig]{Jia21}
Chao Jia, Yinfei Yang, Ye~Xia, Yi-Ting Chen, Zarana Parekh, Hieu Pham, Quoc~V.
  Le, Yunhsuan Sung, Zhen Li, and Tom Duerig.
\newblock Scaling up visual and vision-language representation learning with
  noisy text supervision.
\newblock \emph{arXiv preprint arXiv:2102.05918}, 2021.

\bibitem[Katharopoulos et~al.(2020)Katharopoulos, Vyas, Pappas, and
  Fleuret]{Katharopoulos21Linear}
Angelos Katharopoulos, Apoorv Vyas, Nikolaos Pappas, and Fran\c{c}ois Fleuret.
\newblock Transformers are {RNNs}: Fast autoregressive transformers with linear
  attention.
\newblock In \emph{Proc. ICML}, 2020.

\bibitem[Kay et~al.(2017)Kay, Carreira, Simonyan, Zhang, Hillier,
  Vijayanarasimhan, Viola, Green, Back, Natsev, Suleyman, and Zisserman]{Kay17}
Will Kay, Jo{\~{a}}o Carreira, Karen Simonyan, Brian Zhang, Chloe Hillier,
  Sudheendra Vijayanarasimhan, Fabio Viola, Tim Green, Trevor Back, Paul
  Natsev, Mustafa Suleyman, and Andrew Zisserman.
\newblock The kinetics human action video dataset.
\newblock \emph{arXiv preprint arXiv:1705.06950}, 2017.

\bibitem[Kitaev et~al.(2020)Kitaev, Kaiser, and Levskaya]{kitaev2020reformer}
Nikita Kitaev, Lukasz Kaiser, and Anselm Levskaya.
\newblock Reformer: The efficient transformer.
\newblock In \emph{Proc. ICLR}, 2020.

\bibitem[Korbar et~al.(2019)Korbar, Tran, and Torresani]{Korbar2019scsampler}
Bruno Korbar, Du~Tran, and Lorenzo Torresani.
\newblock Scsampler: Sampling salient clips from video for efficient action
  recognition.
\newblock In \emph{Proc. ICCV}, 2019.

\bibitem[Kuehne et~al.(2011)Kuehne, Jhuang, Garrote, Poggio, and
  Serre]{Kuehne11}
H.~Kuehne, H.~Jhuang, E.~Garrote, T.~Poggio, and T.~Serre.
\newblock {HMDB}: A large video database for human motion recognition.
\newblock In \emph{Proc. ICCV}, 2011.

\bibitem[Kuehne et~al.(2014)Kuehne, Arslan, and Serre]{Kuehne12}
Hilde Kuehne, Ali Arslan, and Thomas Serre.
\newblock The language of actions: Recovering the syntax and semantics of
  goal-directed human activities.
\newblock In \emph{Proc. CVPR}, 2014.

\bibitem[Lample et~al.(2019)Lample, Sablayrolles, Ranzato, Denoyer, and
  J{\'e}gou]{lample2019key}
Guillaume Lample, Alexandre Sablayrolles, Marc'Aurelio Ranzato, Ludovic
  Denoyer, and Herv{\'e} J{\'e}gou.
\newblock Large memory layers with product keys.
\newblock In \emph{NeurIPS}, 2019.

\bibitem[Law and Deng(2018)]{Law18}
Hei Law and Jia Deng.
\newblock {CornerNet}: Detecting objects as paired keypoints.
\newblock In \emph{Proc. ECCV}, 2018.

\bibitem[Lin et~al.(2017)Lin, Goyal, Girshick, He, and Doll{\'{a}}r]{Lin17}
Tsung-Yi Lin, Priya Goyal, Ross Girshick, Kaiming He, and Piotr Doll{\'{a}}r.
\newblock Focal loss for dense object detection.
\newblock In \emph{Proc. ICCV}, 2017.

\bibitem[Lin et~al.(2022)Lin, Petroni, Bertasius, Rohrbach, Chang, and
  Torresani]{Lin22DistSup}
Xudong Lin, Fabio Petroni, Gedas Bertasius, Marcus Rohrbach, Shih-Fu Chang, and
  Lorenzo Torresani.
\newblock Learning to recognize procedural activities with distant supervision.
\newblock In \emph{Proc. CVPR}, 2022.

\bibitem[Liu et~al.(2016)Liu, Anguelov, Erhan, Szegedy, Reed, Fu, and
  Berg]{Liu16}
Wei Liu, Dragomir Anguelov, Dumitru Erhan, Christian Szegedy, Scott Reed,
  Cheng-Yang Fu, and Alexander~C Berg.
\newblock {SSD}: Single shot multibox detector.
\newblock In \emph{Proc. ECCV}, pages 21--37. Springer, 2016.

\bibitem[Loshchilov and Hutter(2017)]{loshchilov2017sgdr}
Ilya Loshchilov and Frank Hutter.
\newblock {SGDR}: Stochastic gradient descent with warm restarts.
\newblock In \emph{Proc. ICLR}, 2017.

\bibitem[Miech et~al.(2019)Miech, Zhukov, Alayrac, Tapaswi, Laptev, and
  Sivic]{Miech19}
Antoine Miech, Dimitri Zhukov, Jean-Baptiste Alayrac, Makarand Tapaswi, Ivan
  Laptev, and Josef Sivic.
\newblock How{T}o100{M}: Learning a text-video embedding by watching hundred
  million narrated video clips.
\newblock In \emph{Proc. ICCV}, 2019.

\bibitem[Miech et~al.(2020)Miech, Alayrac, Smaira, Laptev, Sivic, and
  Zisserman]{Miech20}
Antoine Miech, Jean-Baptiste Alayrac, Lucas Smaira, Ivan Laptev, Josef Sivic,
  and Andrew Zisserman.
\newblock End-to-end learning of visual representations from uncurated
  instructional videos.
\newblock In \emph{Proc. CVPR}, 2020.

\bibitem[Pathak et~al.(2016)Pathak, Kr{\"{a}}henb{\"{u}}hl, Donahue, Darrell,
  and Efros]{Pathak16}
Deepak Pathak, Philipp Kr{\"{a}}henb{\"{u}}hl, Jeff Donahue, Trevor Darrell,
  and Alexei~A. Efros.
\newblock Context encoders: Feature learning by inpainting.
\newblock In \emph{Proc. CVPR}, 2016.

\bibitem[Patrick et~al.(2021{\natexlab{a}})Patrick, Asano, Kuznetsova, Fong,
  Henriques, Zweig, and Vedaldi]{Patrick2020gdt}
Mandela Patrick, Yuki~M. Asano, Polina Kuznetsova, Ruth Fong, João~F.
  Henriques, Geoffrey Zweig, and Andrea Vedaldi.
\newblock On composition of transformations in contrastive self-supervised
  learning.
\newblock In \emph{Proc. ICCV}, 2021{\natexlab{a}}.

\bibitem[Patrick et~al.(2021{\natexlab{b}})Patrick, Campbell, Asano, Misra,
  Metze, Feichtenhofer, Vedaldi, and Henriques]{Patrick21MF}
Mandela Patrick, Dylan Campbell, Yuki~M. Asano, Ishan Misra, Florian Metze,
  Christoph Feichtenhofer, Andrea Vedaldi, and Jo{\~{a}}o~F. Henriques.
\newblock Keeping your eye on the ball: Trajectory attention in video
  transformers.
\newblock In \emph{NeurIPS}, 2021{\natexlab{b}}.

\bibitem[Qian et~al.(2020)Qian, Meng, Gong, Yang, Wang, Belongie, and
  Cui]{Qian20}
Rui Qian, Tianjian Meng, Boqing Gong, Ming-Hsuan Yang, Huisheng Wang, Serge
  Belongie, and Yin Cui.
\newblock Spatiotemporal contrastive video representation learning.
\newblock \emph{arXiv preprint arXiv:2008.03800}, 2020.

\bibitem[Radford et~al.(2021)Radford, Kim, Hallacy, Ramesh, Goh, Agarwal,
  Sastry, Askell, Mishkin, Clark, Krueger, and Sutskever]{Radford21}
Alec Radford, Jong~Wook Kim, Chris Hallacy, Aditya Ramesh, Gabriel Goh,
  Sandhini Agarwal, Girish Sastry, Amanda Askell, Pamela Mishkin, Jack Clark,
  Gretchen Krueger, and Ilya Sutskever.
\newblock Learning transferable visual models from natural language
  supervision.
\newblock \emph{arXiv preprint arXiv:2103.00020}, 2021.

\bibitem[Rastegari et~al.(2016)Rastegari, Ordonez, Redmon, and
  Farhadi]{Rastegari16}
Mohammad Rastegari, Vicente Ordonez, Joseph Redmon, and Ali Farhadi.
\newblock {XNOR-Net}: Imagenet classification using binary convolutional neural
  networks.
\newblock In \emph{Proc. ECCV}, 2016.

\bibitem[Redmon et~al.(2016)Redmon, Divvala, Girshick, and Farhadi]{Redmon16}
Joseph Redmon, Santosh Divvala, Ross Girshick, and Ali Farhadi.
\newblock You only look once: Unified, real-time object detection.
\newblock In \emph{Proc. CVPR}, 2016.

\bibitem[Roy et~al.(2020)Roy, Saffar, Vaswani, and Grangier]{Roy20Routing}
Aurko Roy, Mohammad Saffar, Ashish Vaswani, and David Grangier.
\newblock Efficient content-based sparse attention with routing transformers.
\newblock 2020.

\bibitem[Simonyan and Zisserman(2014)]{Simonyan14b}
Karen Simonyan and Andrew Zisserman.
\newblock Two-stream convolutional networks for action recognition in videos.
\newblock In \emph{NeurIPS}, 2014.

\bibitem[Song et~al.(2020)Song, Tan, Qin, Lu, and Liu]{song2020mpnet}
Kaitao Song, Xu~Tan, Tao Qin, Jianfeng Lu, and Tie-Yan Liu.
\newblock {MPNet}: Masked and permuted pre-training for language understanding.
\newblock \emph{arXiv preprint arXiv:2004.09297}, 2020.

\bibitem[Soomro et~al.(2012)Soomro, Zamir, and Shah]{Soomro12}
Khurram Soomro, Amir~Roshan Zamir, and Mubarak Shah.
\newblock {UCF101}: A dataset of 101 human actions classes from videos in the
  wild.
\newblock \emph{arXiv preprint arXiv:1212.0402}, 2012.

\bibitem[Szegedy et~al.(2016)Szegedy, Vanhoucke, Ioffe, Shlens, and
  Wojna]{Szegedy16}
Christian Szegedy, Vincent Vanhoucke, Sergey Ioffe, Jonathon Shlens, and
  Zbigniew Wojna.
\newblock Rethinking the inception architecture for computer vision.
\newblock In \emph{NeurIPS}, 2016.

\bibitem[Tan and Le(2019)]{Tan2019eff}
Mingxing Tan and Quoc~V. Le.
\newblock Efficientnet: Rethinking model scaling for convolutional neural
  networks.
\newblock In \emph{Proc. ICML}, 2019.

\bibitem[Tang et~al.(2019)Tang, Ding, Rao, Zheng, Zhang, Zhao, Lu, and
  Zhou]{Tang2019coin}
Yansong Tang, Dajun Ding, Yongming Rao, Yu~Zheng, Danyang Zhang, Lili Zhao,
  Jiwen Lu, and Jie Zhou.
\newblock Coin: A large-scale dataset for comprehensive instructional video
  analysis.
\newblock In \emph{Proc. CVPR}, 2019.

\bibitem[Tay et~al.(2020)Tay, Bahri, Yang, Metzler, and Juan]{Tay2020sparse}
Yi~Tay, Dara Bahri, Liu Yang, Donald Metzler, and Da-Cheng Juan.
\newblock Sparse sinkhorn attention.
\newblock \emph{arXiv preprint arXiv:2002.11296}, 2020.

\bibitem[Tong et~al.(2022)Tong, Song, Wang, and Wang]{Tong2022VideoMAE}
Zhan Tong, Yibing Song, Jue Wang, and Limin Wang.
\newblock Videomae: Masked autoencoders are data-efficient learners for
  self-supervised video pre-training.
\newblock \emph{arXiv preprint arXiv:2203.12602}, 2022.

\bibitem[van~den Oord et~al.(2018)van~den Oord, Li, and Vinyals]{Oord18}
A{\"{a}}ron van~den Oord, Yazhe Li, and Oriol Vinyals.
\newblock Representation learning with contrastive predictive coding.
\newblock \emph{arXiv preprint arXiv:1807.03748}, 2018.

\bibitem[Vaswani et~al.(2017)Vaswani, Shazeer, Parmar, Uszkoreit, Jones, Gomez,
  Kaiser, and Polosukhin]{Vaswani17}
Ashish Vaswani, Noam Shazeer, Niki Parmar, Jakob Uszkoreit, Llion Jones,
  Aidan~N. Gomez, Lukasz Kaiser, and Illia Polosukhin.
\newblock Attention is all you need.
\newblock In \emph{NeurIPS}, 2017.

\bibitem[Wang et~al.(2020)Wang, Li, Khabsa, Fang, and Ma]{wang2020linformer}
Sinong Wang, Belinda~Z. Li, Madian Khabsa, Han Fang, and Hao Ma.
\newblock Linformer: Self-attention with linear complexity.
\newblock \emph{arXiv preprint arXiv:2006.04768}, 2020.

\bibitem[Wu et~al.(2018)Wu, Zaheer, Hu, Manmatha, Smola, and
  Kr{\"{a}}henb{\"{u}}hl]{Wu18}
Chao-Yuan Wu, Manzil Zaheer, Hexiang Hu, R.~Manmatha, Alexander~J. Smola, and
  Philipp Kr{\"{a}}henb{\"{u}}hl.
\newblock Compressed video action recognition.
\newblock In \emph{Proc. CVPR}, 2018.

\bibitem[Wu et~al.(2020)Wu, Girshick, He, Feichtenhofer, and
  Kr{\"{a}}henb{\"{u}}hl]{Wu20MultiGrid}
Chao-Yuan Wu, Ross Girshick, Kaiming He, Christoph Feichtenhofer, and Philipp
  Kr{\"{a}}henb{\"{u}}hl.
\newblock A multigrid method for efficiently training video models.
\newblock In \emph{Proc. CVPR}, 2020.

\bibitem[Yun et~al.(2019)Yun, Han, Oh, Chun, Choe, and Yoo]{Yun2019Cutmix}
Sangdoo Yun, Dongyoon Han, Seong~Joon Oh, Sanghyuk Chun, Junsuk Choe, and
  Youngjoon Yoo.
\newblock Cutmix: Regularization strategy to train strong classifiers with
  localizable features.
\newblock In \emph{Proc. ICCV}, 2019.

\bibitem[Zhang et~al.(2018{\natexlab{a}})Zhang, Cisse, Dauphin, and
  Lopez-Paz]{zhang2018mixup}
Hongyi Zhang, Moustapha Cisse, Yann~N. Dauphin, and David Lopez-Paz.
\newblock mixup: Beyond empirical risk minimization.
\newblock In \emph{Proc. ICLR}, 2018{\natexlab{a}}.

\bibitem[Zhang et~al.(2018{\natexlab{b}})Zhang, Zhou, Lin, , and
  Sun]{Zhang18-shufflenet}
Xiangyu Zhang, Xinyu Zhou, Mengxiao Lin, , and Jian Sun.
\newblock Shufflenet: An extremely efficient convolutional neural network for
  mobile devices.
\newblock In \emph{Proc. CVPR}, 2018{\natexlab{b}}.

\bibitem[Zhenda et~al.(2022)Zhenda, Zhang, Cao, Lin, Bao, Yao, Dai, and
  Hu]{Xie22SimMiM}
Zhenda, Zheng Zhang, Yue Cao, Yutong Lin, Jianmin Bao, Zhuliang Yao, Qi~Dai,
  and Han Hu.
\newblock {SimMIM}: A simple framework for masked image modeling.
\newblock In \emph{Proc. CVPR}, 2022.

\bibitem[Zhou et~al.(2021)Zhou, Lin, Li, and Zheng]{Zhou21GHRM}
Jiaming Zhou, Kun-Yu Lin, Haoxin Li, and Wei-Shi Zheng.
\newblock Graph-based high-order relation modeling for long-term action
  recognition.
\newblock In \emph{Proc. CVPR}, 2021.

\end{thebibliography}
